\documentclass[letterpaper, 10 pt, conference]{ieeeconf}  

\IEEEoverridecommandlockouts                              

\usepackage{graphics} 
\usepackage{epsfig} 
\usepackage{times} 
\usepackage{amsmath} 
\usepackage{amssymb}  
\usepackage[makeroom]{cancel}

\usepackage{geometry}
\usepackage{shortcuts}

\title{\LARGE \bf
Robust Bipedal Locomotion: \\
Leveraging Saltation Matrices for Gait Optimization} 

\author{Maegan Tucker$^{1}$, Noel Csomay-Shanklin$^{2}$, and Aaron D. Ames$^{1,2}$
\thanks{This research was supported by NSF Graduate Research Fellowship No. DGE‐1745301 and the Zeitlin Family Fund. Research involving human subjects was conducted under IRB No. 21-0693}%
\thanks{$^{1}$ Authors are with the Department
of Mechanical and Civil Engineering, California Institute of Technology,
Pasadena, CA 91125.}%
\thanks{$^{2}$ Authors are with the Department of Control and Dynamical Systems, California Institute of Technology,
Pasadena, CA 91125.}
\thanks{\texttt{mtucker@caltech.edu}}%
}

\begin{document}

\maketitle
\thispagestyle{empty}
\pagestyle{empty}

\begin{abstract}
The ability to generate robust walking gaits on bipedal robots is key to their successful realization on hardware. 
To this end, this work extends the method of Hybrid Zero Dynamics (HZD) -- which traditionally only accounts for locomotive stability via periodicity constraints under perfect impact events -- through the inclusion of the saltation matrix with a view toward synthesizing robust walking gaits. By jointly minimizing the norm of the extended saltation matrix and the torque of the robot directly in the gait generation process, we demonstrate that the synthesized gaits are more robust than gaits generated with either term alone; these results are shown in simulation and on hardware for the AMBER-3M planar biped and the Atalante lower-body exoskeleton (both with and without a human subject). The end result is experimental validation that combining saltation matrices with HZD methods produces more robust bipedal walking in practice. 
\end{abstract}


\section{Introduction}


Achieving stable and robust locomotion on legged systems is a challenging control task due to underactuation, power limitations, and ground impacts. Two main approaches that have proven successful towards mitigating these challenges in the real world include: 1) generating stable reference trajectories \cite{paredes2020dynamic, castillo2020velocity, reher2020algorithmic, paredes2022resolved} and modifying these behaviors online using regulators (such as modifying the swing foot location based on lateral velocity \cite{raibert1984hopping,reher2021dynamic}); and 2) determining the desired behavior of the robot in real time using online planning via model predictive control \cite{kim2019highly, grandia2019feedback, sleiman2021unified, garcia2021mpc} or reinforcement learning \cite{fankhauser2013reinforcement, li2021reinforcement, siekmann2021blind, rudin2022learning}. 
In this work, we aim to improve these existing approaches by synthesizing \textit{robust} reference trajectories. This is motivated by previous work, which has shown that optimizing the robustness of nominal trajectories improves overall performance regardless of the chosen method of online stabilization \cite{mombaur2001stability, dai2012optimizing}, and that online planning strategies can have unpredictable behavior without the use of a reference trajectory \cite{chevallereau2008stable, xie2018feedback}.

\begin{figure}[tb]
    \centering
    \href{https://youtu.be/BZu-9UStG2E}{
    \includegraphics[width=\linewidth]{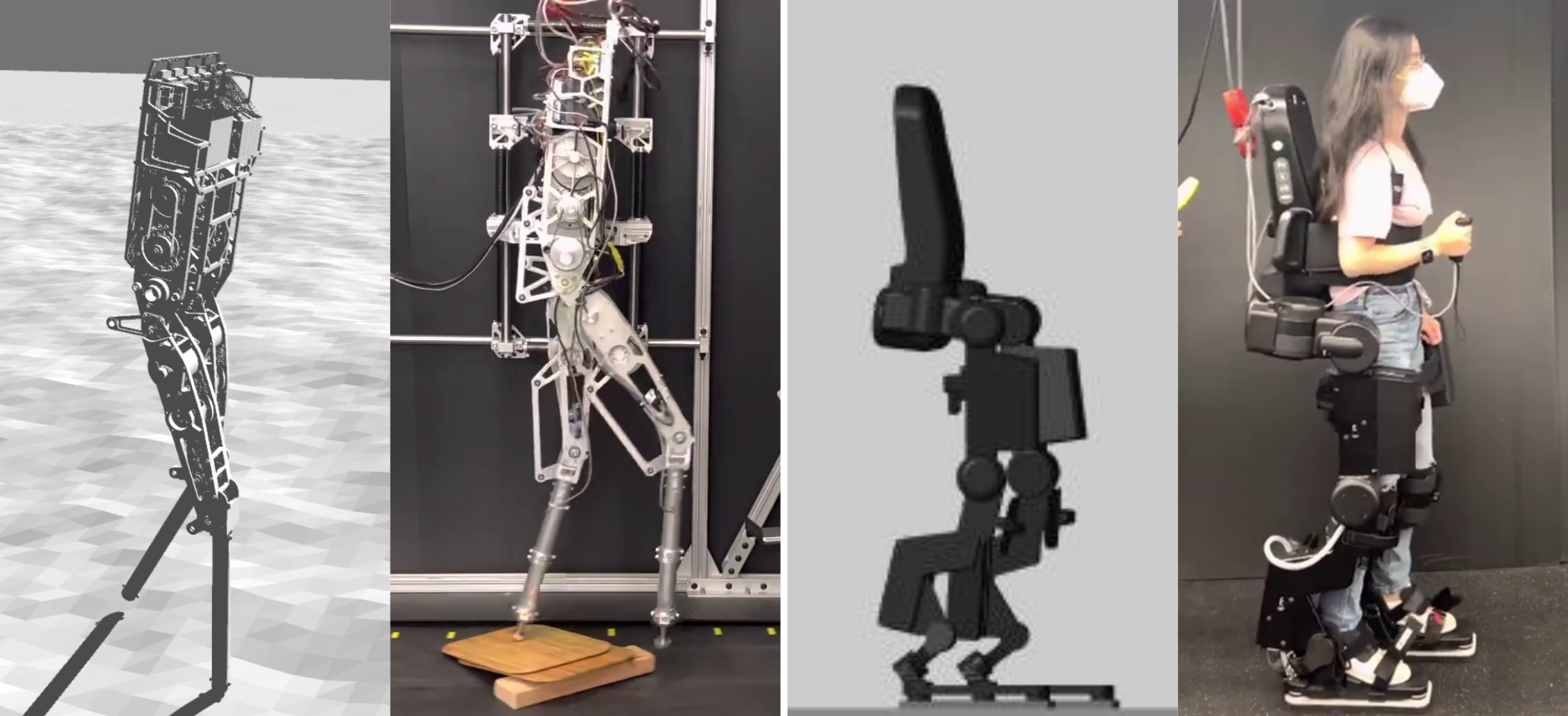}}
    \caption{This work improves the robustness of nominal reference trajectories (gaits) by evaluating the extended saltation matrix directly in the gait generation framework. The approach is demonstrated in simulation and on hardware for both the 7-DOF AMBER-3M planar biped (left) and the 18-DOF Atalante lower-body exoskeleton (right).\vspace{-3mm}}
    \label{fig:intro}
\end{figure}

It is important to note that there exists previous work towards generating robust limit cycles \cite{dai2012optimizing}. However, these existing methods can be computationally expensive and do not scale easily to high-dimensional systems. Thus, the goal of this work is to develop a method of generating robust limit cycles in a way that is scalable to high-dimensional systems (such as the 18 degree-of-freedom (DOF) Atalante lower-body exoskeleton shown in Fig. \ref{fig:intro}).

Our approach for generating robust walking gaits builds upon the Hybrid Zero Dynamics (HZD) method \cite{westervelt2003hybrid, grizzle20103d}. 
Importantly, this approach has been successfully deployed on several high-dimensional real-world legged robots \cite{sreenath2011compliant, ma2017bipedal, hereid2018dynamic, harib2018feedback, ma2019first, reher2020algorithmic}. The HZD framework encodes walking via nontrivial periodic orbits and directly accounts for discrete impact events. One advantage of this approach is that the stability of the walking can be analyzed using the method of Poincar\'e sections for systems with impulse effects \cite{grizzle2001asymptotically}. 
However, in practice, it can be difficult to synthesize and control these periodic orbits since 1) it can be computationally expensive to evaluate the Poincar\'e section for high-dimensional systems and 2) the timing of impact events for real-world systems is noisy, throwing the system off the nominal periodic orbit. For these reasons, the existing approach towards generating robust gaits necessitates extensive user tuning \cite{tucker2021preference}. 


This paper proposes a robust HZD framework that leverages saltation matrices -- first-order approximations of a system's sensitivity to discrete events -- to directly synthesize robust nominal walking trajectories. These matrices, originally used in the field of non-smooth analysis, have been receiving growing attention and have been recently demonstrated towards state estimation for hybrid systems \cite{kong2021salted, payne2022uncertainty} and hybrid event shaping \cite{zhu2022hybrid}. Inspired by this recent research, our work similarly utilizes saltation matrices to generate stable periodic walking gaits that lead to robust behavior in the real world, even for high-dimensional systems; specifically, we propose including the induced norm of the extended saltation matrix in the HZD optimization cost function. As previewed in Fig. \ref{fig:intro}, we demonstrate the proposed framework in simulation and on hardware for the 7-DOF AMBER-3M planar biped and the 18-DOF Atalante lower-body exoskeleton with and without a human subject.


\section{Preliminaries on Gait Generation via the Hybrid Zero Dynamics Method}
\label{sec: prelimHZD}

As walking is comprised of alternating sequences of continuous dynamics followed by intermittent discrete impact events, it is naturally modeled as a hybrid system \cite{lygeros1999existence, lygeros2003dynamical, ames2009three}. Consider a robotic system with coordinates $q \in \mathcal{Q}\subset\mathbb{R}^n$ and system state $x=(q,\dot q)\in \mathsf T\mathcal{Q}$, where $\mathsf T\mathcal{Q}$ denotes the tangent bundle of the configuration manifold $\mathcal{Q}$. To define the hybrid dynamics, we begin by introducing the admissible domain $\mathcal{D} \subset \mathsf T\mathcal{Q}$ on which the system evolves, i.e. the set of states with swing foot above the floor, which respecting joint limits, etc. During the continuous domain, the dynamics can be derived from the Euler-Lagrange equations as:
\begin{align}
    \dot x = \underbrace{\begin{bmatrix}
        \dot q \\ -D(q)^{-1}H(q, \dot q) 
    \end{bmatrix}}_{f(x)} + \underbrace{\begin{bmatrix}0\\D(q)^{-1}B\end{bmatrix}}_{g(x)}u.
\end{align}
where $D:\mathcal{Q}\to \mathbb{R}^{n\times n}$ is the mass-inertia matrix, $H:\mathsf T\mathcal{Q}\to \mathbb{R}^n$ contains the Coriolis and gravity terms, $B\in\mathbb{R}^{n\times m}$ is the actuation matrix, and $u \in \R^m$ is the control input. 

Next, we define the \textit{guard}, the set of states where a discrete impact event will occur. Letting $p_{sw}^z:\mathcal{Q}\to\mathbb{R}$ return the vertical position of the swing foot, the guard for the systems investigated in this work (assuming a known and constant ground height of zero) is defined as: \begin{align}
    \mathcal{S} = \{x\in \mathcal{D} ~|~ p^z_{sw}(q) = 0,~ \dot p^z_{sw}(x) < 0\},
\end{align}
or the set of states when the swing foot strikes the ground with a negative velocity. As the system flows into the guard, $n_c \in \mathbb{N}$ holonomic constraints $c:\mathcal{Q}\to \R^{n_c}$ are enforced in the continuous domain succeeding the impact event, and the system undergoes a discrete jump in the state as captured by the momentum transfer equation \cite{hurmuzlu1994rigid}:
\begin{align}
    D(q^-)(\dot q^+ - \dot q^-) = J_c(q^-)^\top \delta F
\end{align}
where $q^-, q^+ \in \mathcal{Q}$, and $\dot q^-, \dot q^+\in\R^n$ represent the robot configuration and velocity just before and after impact, respectively, $\delta F \in \R^{n_c}$ is the impulse force of the impact event, and $J_c:\mathcal{Q}\to \R^{n_c\times n}$ is the Jacobian of the holonomic constraints.


Enforcing the holonomic constraints through impacts can equivalently be represented as $J_c(q^-)\dot q^+ = 0$. With this, we can rewrite the impact equation \cite{glocker1992dynamical} as:
\begin{align}
    \begin{bmatrix}
        D(q^-) & -J_c(q^-)^\top \\ J_c(q^-) & 0
    \end{bmatrix}
    \begin{bmatrix}
        \dot{q}^+ \\ \delta F
    \end{bmatrix} = 
    \begin{bmatrix}
        D(q^-)\dot{q}^{-} \\ 0
    \end{bmatrix}. \label{eq: impactrelation}
\end{align}
Noting that the configuration is continuous through impact, solving for $\dot q^+$ in \eqref{eq: impactrelation} allows us to define the \textit{reset map} $\Delta: \mathcal{S} \to \mathcal{D}$, i.e. the map describing the discrete event at footstrike \cite{grizzle2014models}, as:
\begin{align}
    \small
    x^+ = \Delta(x^-) := 
    \begin{bmatrix}
        Rq^- \\ R(-D^{-1}J_c^{\top}(J_c D^{-1}J_c^{\top})^{-1} J_c + I)\dot{q}^-    
    \end{bmatrix}
    \label{eq: reset}
\end{align}
where the dependence of $D$ and $J_c$ on $q^-$ is suppressed and $R \in \R^{n \times n}$ denotes the relabeling matrix which is used to maintain state consistency between domains. We are now fully equipped to define the hybrid system of walking as:
\begin{align}
\mathcal{H}\mathcal{C} = 
    \begin{cases}
        \dot{x} = f(x) + g(x)u, & x \in \mathcal{D} \backslash\mathcal{S}, \\
        x^+ = \Delta (x^-), & x \in \mathcal{S}.
    \end{cases}
\end{align}
Note that hybrid systems with a more diverse collection of contact sequences can be modeled via the same framework \cite{sinnet20092d, zhao2017multi, reher2020algorithmic}, at the cost of introducing a directed graph describing how the continuous and discrete domains are related -- this is omitted in this work for the sake of simplicity.

\newsec{Trajectory Optimization}
The hybrid zero dynamics (HZD) method of gait generation leverages trajectory optimization in the context of the aforementioned hybrid control system to synthesize provably stable walking trajectories (gaits), encoded as nontrivial limit cycles \cite{grizzle20103d}.  

 We begin by defining a collection of $k \in \mathbb{N}$ \textit{outputs} or \textit{virtual constraints} $y_{\alpha}: \mathcal{Q} \to \R^k$ which we would like to converge to zero. These virtual constraints encode the desired behavior of the system via:
\begin{align}
    y_{\alpha}(q) = y^a(q) - y^d_\alpha(\tau(q)),
\end{align}
where $y^a: \mathcal{Q} \to \R^k$ represents the actual (measured) outputs of the system, $y^d_\alpha: \R \to \R^k$ represents the desired outputs commonly parameterized via a $p^{\text{th}}$-order B\'ezier polynomial with B\'ezier coefficients $\alpha \in \R^{k \times p+1}$, and $\tau: \mathcal{Q} \to [0,1]$ is a monotonically increasing variable over the gait cycle, termed a \textit{phasing variable}. The HZD framework reduces the stability analysis of the system $\mathcal{H}\mathcal{C}$ to a lower-dimensional manifold, the \textit{zero dynamics surface}:
\begin{align}
    \mathcal{Z}_{\alpha} := \{ x \in \mathcal{D} ~|~ y_{\alpha}(q) = 0, \dot{y}_{\alpha}(x) = 0\},
\end{align}
%
%
%
%
which can be rendered impact-invariant by enforcing the \textit{HZD condition}:
\begin{align}
    \Delta(\mathcal{S} \cap \mathcal{Z}_{\alpha}) \subset \mathcal{Z}_{\alpha}. \label{eq: HZDcondition}
\end{align}

Driving the outputs to zero using a stabilizing controller $u^*: \mathcal{D} \to \R^m$, for example a feedback linearizing or control Lyapunov function based controller, results in a closed loop dynamical system: $\dot{x} = f_{cl}(x) := f(x) + g(x)u^*(x)$. 
This stabilizing controller, paired with the HZD condition and an orbit which is stable on the zero dynamics surface renders the closed loop hybrid dynamical system stable. 
Importantly, the stability of the zero dynamics on $\mathcal{Z}_{\alpha}$ can be shaped through the choice of outputs and B\'ezier coefficients $\alpha$. Therefore, an optimization problem is constructed to synthesize trajectories with desired outputs such that \eqref{eq: HZDcondition} is enforced. This optimization problem takes the form:
\begin{align}
    \{\alpha^*, X^*\} &= \argmin_{\alpha, X} ~\Phi(X) \label{eq: NLP}\\
    \text{s.t.} \quad &\dot{x} = f_{cl}(x) \tag{Closed-loop Dynamics} \\
    &\Delta(\mathcal{S} \cap \mathcal{Z}_{\alpha}) \subset \mathcal{Z}_{\alpha} \tag{HZD Condition} \\
    &X_{\text{min}} \preceq X \preceq X_{\text{max}} \tag{Decision Variables} \\
    &a_{\text{min}} \preceq a(X) \preceq a_{\text{max}} \tag{Physical Constraints}
\end{align}
In this NLP, $X = (x_0, \dots, x_N, T) \in \mathcal{X} $ is the collection of all decision variables with $x_i \in \R^{2n}$ being the state at the $i^{\text{th}}$ collocation point and $T \in \R$ the total duration, $\Phi: \mathcal{X} \to \R$ denotes the cost function (such as torque-squared or mechanical cost of transport \cite{reher2020algorithmic}), and $a(X)$ is the set of physical constraints which includes holonomic constraints, workspace limits, power limits, etc. The solution of \eqref{eq: NLP} is a limit cycle which encodes stable walking, described by some static set of B\'ezier coefficients $\alpha^* \in \R^{k \times p+1}$.


\section{Preliminaries on Saltation Matrices}
\label{sec: saltation}

\begin{figure}[tb]
    \centering
    \includegraphics[width=\linewidth]{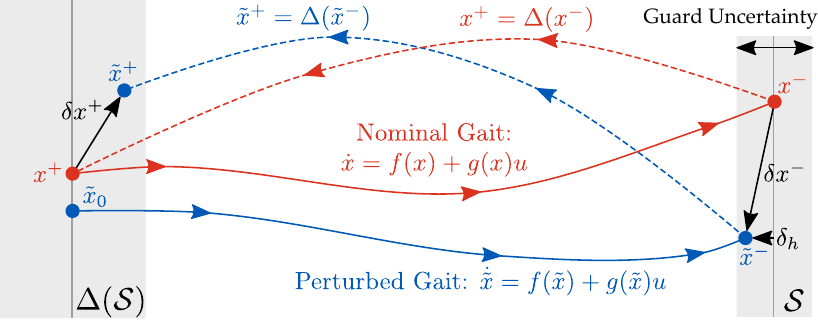}
    \caption{Illustration of a perturbed flow (blue) with uncertain guard conditions (illustrated by the grey region) compared to a nominal periodic orbit (red) which assumes a known guard (illustrated by the black vertical line). The perturbed initial condition $\tilde{x}_0 \in \Delta(\S)$ results in pre-impact state error $\delta x^-$ and post-impact state error $\delta x^+$. In general, saltation matrices capture the relationship between these errors: $\delta x^+ = S \delta x^-$.}
    \label{fig:saltation}
    \vspace{-2mm}
\end{figure}

In preparation for introducing the proposed \textit{robust} HZD method, we will first present some preliminary information on saltation matrices. The saltation matrix is a standard tool used in the field of non-smooth analysis that describes a systems sensitivity to discontinuities (otherwise called `saltations' or `jumps') \cite{leine2013dynamics}. Typically, the saltation matrix $S({t}_i, x({t_i})) \in \R^{2n\times 2n}$ is defined at time ${t}_i \in \R^+$ for states in the guard $x({t}_i) \in \S$, by the relationship:
\begin{align}
    \delta x({t}_{i+1}) = S({t}_i,x({t}_i)) \delta x({t}_i), \label{eq:saltdeftraditional}
\end{align}
where $x(t_{i+1}) \in \Delta(\S)$ denotes the post-transition state. Since we are interested in evaluating the saltation matrix for pre-computed gaits with a known impact time $T_I \in \R^+$, we will specifically derive the saltation matrix $S := S(T_I, x^-)$ for the relationship:
\begin{align}
    \delta x^+ = S \delta x^-, \label{eq:saltdef}
\end{align}
where $\delta x^-, \delta x^+ \in \R^{2n}$ define the variation in the pre- and post- impact states, respectively, at the time of impact $T_I$.
A visualization of these variations are provided in Fig. \ref{fig:saltation}. Explicitly, these state variations are defined as:
\begin{align}
    \delta x^+ := \Tilde{x}^+ - x^+, \quad \delta x^- := \Tilde{x}^- - x^-,
\label{eq:deltaxplus}
\end{align}
where $\Tilde{x}^-,\Tilde{x}^+ \in \R^{2n}$ denote the perturbed states. Let $F^-,F^+ \in \R^{2n}$ respectively denote the vector fields of the linearized closed-loop dynamics evaluated at the pre-impact state $x^-$ and post-impact state $x^+$, defined as:
\begin{align}
    F^- := f_{cl}(x^-), ~F^+ := f_{cl}(x^+).
\end{align}

As explained in \cite{leine2013dynamics}, the derivation of the saltation matrix approximates the perturbed states by flowing the system along $F^-,F^+$ for some duration of time $\delta t \in \R$. Specifically, without loss of generality, the post-impact state can be represented as:
\begin{align}
    x^+ = \Delta(x^-) + F^+ \delta t,
\end{align}
and the perturbed states as:
\begin{align}
    \Tilde{x}^+ &:= \Delta(\tilde{x}^-), \\
    \Tilde{x}^- &:= x^- + \delta x^- + F^- \delta t.
\end{align}
By substituting these approximations into \eqref{eq:deltaxplus} and taking the first-order Taylor series expansion, we obtain:
\begin{align}
    \delta x^+ &= \Delta(x^- + \delta x^- + F^- \delta t) - \Delta(x^-) - F^+ \delta t, \\
    &\approx J_{\Delta}\delta x^- + J_{\Delta} F^- \delta t - F^+ \delta t.  \label{eq:deltaxplus2}
\end{align}
Here, $J_{\Delta} := \frac{\partial }{\partial x} \Delta(x^-) \in \R^{2n \times 2n}$ denotes the Jacobian of the reset map evaluated at the pre-impact state $x^-$. Note that because this work exclusively utilizes the reset map shown in \eqref{eq: reset}, we continue the derivation using the assumption that the reset map has no dependence on time, i.e. $\frac{\partial \Delta}{\partial t} = 0_{2n \times 2n}$. 

Lastly, we can represent $\delta t$ in terms of the pre-impact state error $\delta x^-$ by observing the perturbed guard condition. For generality, we will denote the guard condition as $h: \R^{2n} \to \R$, but note that in our work we specifically define $h(x) := p_{sw}^z(q)$. Applying a first-order Taylor series expansion to the perturbed guard condition, we obtain:
\begin{align}
    0 &= h(\Tilde{x}^-), \label{eq: guardassump} \\
      &\approx \underbrace{h(x^-)}_{=0} + J_h^{\top}(\delta x^- + F^- \delta t). \label{eq: guardassump2}
\end{align}
Here, $J_h :=\frac{\partial}{\partial x}h(x^-) \in \R^{2n}$ denotes the Jacobian of the guard condition with respect to $x^-$. Similar to before, we assume that the guard condition has no dependence on time, i.e. $\frac{\partial h}{\partial t} = 0$; such reset maps are termed \textit{autonomous switching boundary functions}. For information on deriving the saltation matrix for non-autonomous switching boundaries (i.e. $h(x,t)$) refer to \cite{leine2013dynamics}. Through the manipulation of \eqref{eq: guardassump2}, we arrive at our expression of $\delta t$ in terms of $\delta x^-$:
\begin{align}
\delta t &= \frac{-J_h^{\top} \delta x^-}{J_h^{\top} F^-}.
\label{eq:dt}
\end{align}
Using this relationship, we can substitute \eqref{eq:dt} into \eqref{eq:deltaxplus2} to obtain the ``traditional saltation matrix'', $S \in \R^{2n \times 2n}$: 
\begin{align}
    \delta x^+ := \underbrace{\left(
    J_{\Delta} + \frac{(F^+ - J_{\Delta}F^-) J_h^{\top} }{J_h^{\top} F^-},\right)}_{S}\delta x^-. \label{eq: tradsalt}
\end{align}

\newsec{Accounting for Guard Uncertainty}
Recently, Payne et al. \cite{payne2022uncertainty} extended the traditional saltation matrix to also account for guard uncertainty. This is accomplished by adapting \eqref{eq: guardassump} to also account for perturbations in the guard location along the normal direction, denoted as $\delta_h \in \R$ (shown in Fig. \ref{fig:saltation}):
\begin{align}
    \delta_h &= h(\tilde{x}^-) \\
    \delta_h &\approx J_h^{\top} (\delta x^- + F^- \delta t) \\
    \delta t &= \frac{-J_h^{\top} \delta x^- + \delta_h}{J_h^{\top}F^-}. \label{eq: dt2}
\end{align}
Then, substituting \eqref{eq: dt2} into \eqref{eq:deltaxplus2}, we arrive at the expression:
\begin{align}
    \delta x^+ = S \delta x^- + \underbrace{\left(\frac{J_{\Delta}F^- - F^+}{J_h^{\top}F^-}\right)}_{S_g} \delta_h,
\end{align}
where $S$ is the same as in \eqref{eq: tradsalt}, and $S_g \in \R^{2n \times 1}$ is termed the \textit{guard saltation matrix}. Together, these matrices can be combined together to obtain the \textit{extended saltation matrix}, $S_e \in \R^{2n+1 \times 2n+1}$, which is defined by the expression:
\begin{align}
    \begin{bmatrix}
        \delta x^+ \\ \delta_h 
    \end{bmatrix} = 
    \underbrace{\begin{bmatrix}
        S & S_g \\
        0 & 1
    \end{bmatrix}}_{S_e}
    \begin{bmatrix}
        \delta x^- \\ \delta_h 
    \end{bmatrix}. \label{eq: extendedsalt}
\end{align}

\section{Robust Gait Generation}
\label{sec: method}

Now that we have presented the preliminaries on the Hybrid Zero Dynamics (HZD) method and saltation matrices, we will discuss how and why we incorporate the saltation matrix evaluation in the HZD framework. 
First, to address the question of \textit{why} evaluating the saltation matrix improves the robustness of generated gaits, we refer to the field of contraction theory. As first noted by Lohmiller in 1988, discussing stability alone does not capture the behavior of a system relative to a nominal motion \cite{lohmiller1998contraction}. Instead, Lohmiller proposed a new field of analysis -- contraction analysis -- which explores how trajectories evolve relative to nearby trajectories. Specifically, a system is defined as \textit{contractive} if all trajectories converge to some nominal trajectory. In this work, we leverage the notion of contractivity to define \textit{robust} walking behaviors as those that are more contractive. In other words, when disturbed, a robust gait will converge to a nontrivial periodic orbit faster than a non-robust gait.

Moreover, Burden et al. \cite{burden2018contraction} recently leveraged saltation matrices to induce contractivity of a hybrid system through discrete events. Motivated by this, we propose optimizing the contractivity of discrete events by minimizing the induced matrix norm of the extended saltation matrix; the induced matrix norm is equivalent to the largest singular value of the extended saltation matrix (i.e. $\|S_e\|_2 := \sqrt{\lambda_{\text{max}}(S_e^{\top}S_e)} \in \R$). Lastly, since the HZD gait generation framework relies heavily on the cost function, we propose adding the induced norm of the extended saltation matrix directly to the cost.

\begin{figure}
    \centering
    \includegraphics[width=\linewidth]{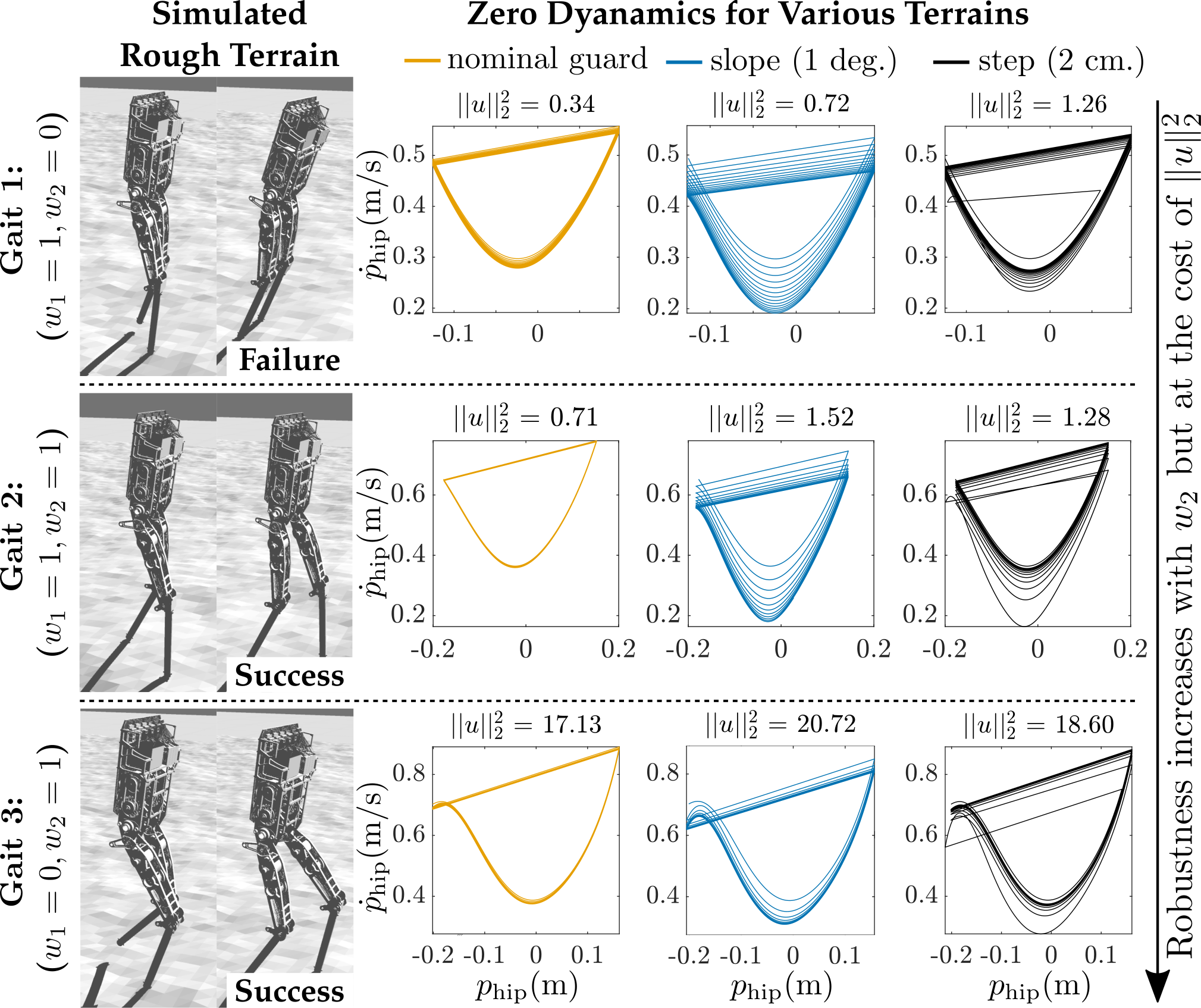}
    \caption{This figure illustrates the simulated behavior of three gaits for AMBER-3M, where each gait was generated with different weighting terms $w_1$ and $w_2$. The behavior is illustrated using the zero dynamics coordinates of the linearized hip position and velocity ($p_{\textrm{hip}}, \dot p_{\textrm{hip}} \in \R$), across three different environment conditions: flat ground as captured by the nominal guard (left); 1 degree slope (middle); and 2cm step height (right). The phase portraits show that the robustness of the walking behavior increases as $w_2$ increases relative to $w_1$, but the gait with $w_1 = 0$ results in significantly increased torque.}
    \label{fig: ambersimgaits}
    \vspace{-4mm}
\end{figure}


\begin{figure*}[tb]
    \centering
    \includegraphics[width=\linewidth]{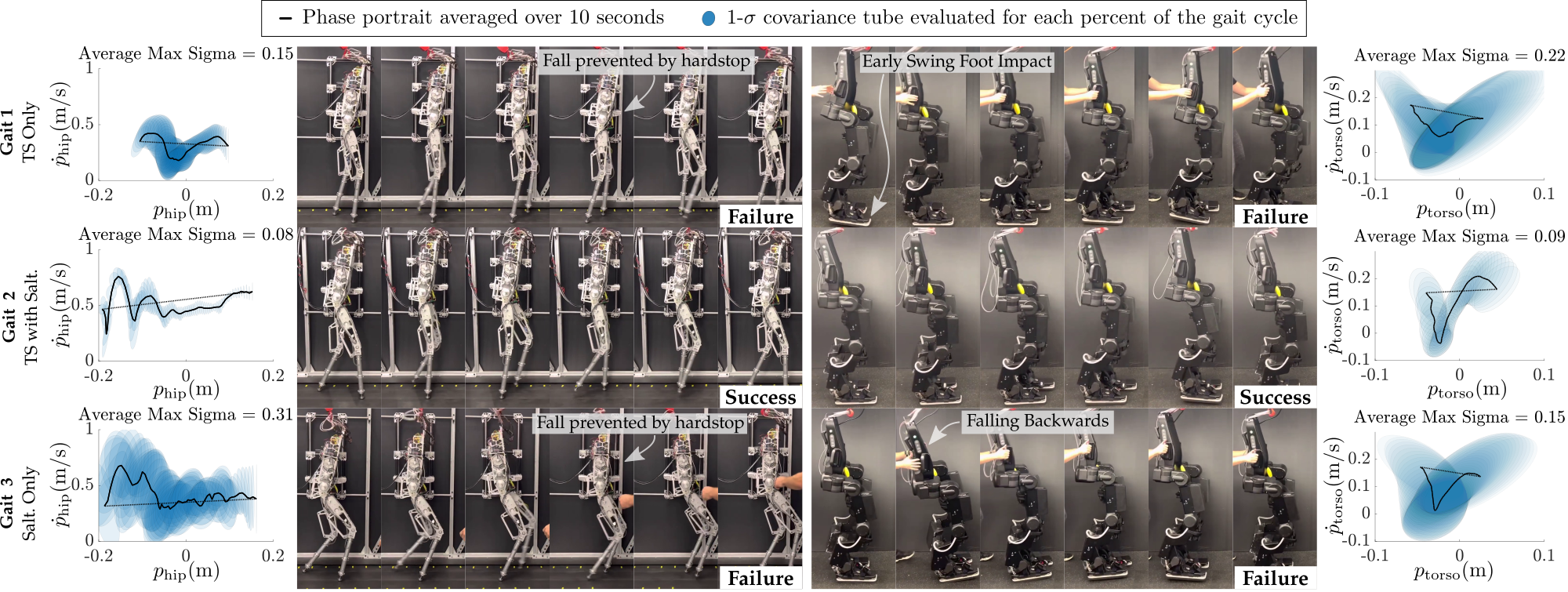}
    \caption{Gait tiles demonstrating the experimental performance of all three gaits on the AMBER-3M planar biped (left), and the empty Atalante exoskeleton (right). For both platforms, only the gait generated with both torque and the saltation matrix in the cost function ($w_1,w_2 > 0$) was able to sustain stable locomotion. The experimental data is also visualized via phase diagrams of the linearized hip position and velocity ($p_{\textup{hip}},\dot p_{\textup{hip}} \in \R$) for AMBER-3M and the forward position and velocity of the floating-base frame ($p_{\textup{torso}},\dot p_{\textup{torso}} \in \R$) relative to the stance foot for the exoskeleton. The black line shows the average zero dynamics across a single step, with the blue region illustrating the 1-sigma tube.}
    \label{fig:gaittiles}
    \vspace{-2mm}
\end{figure*}

To investigate the influence of including the saltation matrix in the optimization problem, the remainder of this paper compares gaits generated with different weightings of the commonly used cost function torque-squared and the induced matrix norm of the extended saltation matrix:
\begin{align}
    \Phi(X) = w_1 \|U\|_2^2 + w_2 \| S_e \|^2_2. \label{eq: costfunc}
\end{align} 
Here, $w_1,w_2 \in \R_{\geq 0}$ denote constant weighting terms, $U \in \R^{m \times N}$ denotes the vectorized torques throughout the nominal gait (assuming a decision variable with $N \in \R$ discretizations) and $S_e \in \R^{2n+1 \times 2n+1}$ denotes the extended saltation matrix evaluated at the pre-impact state of the generated nominal gait. The extended saltation matrix is again computed as in \eqref{eq: extendedsalt}, with the traditional saltation matrix and guard saltation matrix explicitly computed as:
\begin{align*}
    S = J_{\Delta} + \frac{(\dot{x}^+ - J_{\Delta}\dot{x}^-) J_h^{\top} }{J_h^{\top} \dot{x}^-},  ~S_g = \frac{J_{\Delta}\dot{x}^- - \dot{x}^+}{J_h^{\top} \dot{x}^-}.
\end{align*}
To preview the results presented in Sec. \ref{sec: results}, it was found that increasing $w_2$ in \eqref{eq: costfunc} relative to $w_1$ improves robustness of the generated gaits, but also results in increased torque (this is illustrated in Fig. \ref{fig: ambersimgaits}). This trade-off between performance (characterized by successful implementation on hardware) and robustness (characterized by a systems ability to return to nominal periodic orbits in the presence of disturbances) is further explored in the experimental results.

Lastly, we would like to note that computing the Jacobian of the reset map, $J_{\Delta}$, can be computationally expensive for high-dimensional systems because the reset map \eqref{eq: reset} requires several matrix inversions. Thus, for implementation purposes, we numerically approximate the Jacobian of the reset map. However, future work could more efficiently obtain these terms using autodiff or other tools for computing efficient analytical derivatives \cite{singh2022efficient}.
\section{Results}
\label{sec: results}

We demonstrate the application of saltation matrices towards robust gait generation on two robotic platforms: the AMBER-3M planar biped, and the Atalante exoskeleton. As illustrated in Fig. \ref{fig: ambersimgaits}, three gaits were generated for each robotic platform: 1) a nominal gait with the cost function equal to torque squared; 2) a gait with the cost function being a weighting of both torque squared and the induced matrix norm of the extended saltation matrix at impact; and 3) a gait with the cost function only including the saltation matrix. For AMBER-3M, the three compared gaits were generated with weight values of\footnote{The purpose of scaling $w_1$ and $w_2$ is to ensure that the platform-specific torque-squared term $\|U\|_2^2$ has a similar magnitude to that of $\|S_e\|_2^2$, thus preventing either term from heavily dominating the NLP objective.}: $w_1 = 1, w_2 = 0$; $w_1 = 1, w_2 = 1$; $w_1 = 0, w_2 = 1$. For Atalante, the weights were selected as: $w_1 = 1e^{-6}, w_2 = 0$; $w_1 = 1e^{-6}, w_2 = 1e^{6}$; $w_1 = 0, w_2 = 1e^{6}$. Note that for each robotic platform, the remaining constraints and bounds of the HZD gait generation framework were held constant; only the weights $w_1$ and $w_2$ varied. All gaits were generated using the FROST toolbox \cite{hereid2017frost}. The presented experimental results are best demonstrated via the supplemental video \cite{video}.

\newsec{AMBER-3M Results}
The AMBER-3M planar biped\footnote{In this work, we specifically utilize the point-foot configuration of AMBER-3M, termed AMBER3M-PF.} is a custom planar robot \cite{ambrose2017toward} with four motorized joints (left hip, left knee, right hip, right knee). The measured joint positions are denoted as $q^a \in \R^4$ and are selected as the outputs of the generated gaits, i.e $y^a(q) := q^a \in \R^4$. The phasing variable $\tau(q) \in \R$ is selected to be the linearized forward hip position, and the desired outputs are described using a $5^{\text{th}}$-order B\'ezier polynomial ($\alpha \in \R^{4 \times 6}$). The generated joint-level trajectories are enforced using PD control.
%


\begin{figure*}
    \centering
    \includegraphics[width=\linewidth]{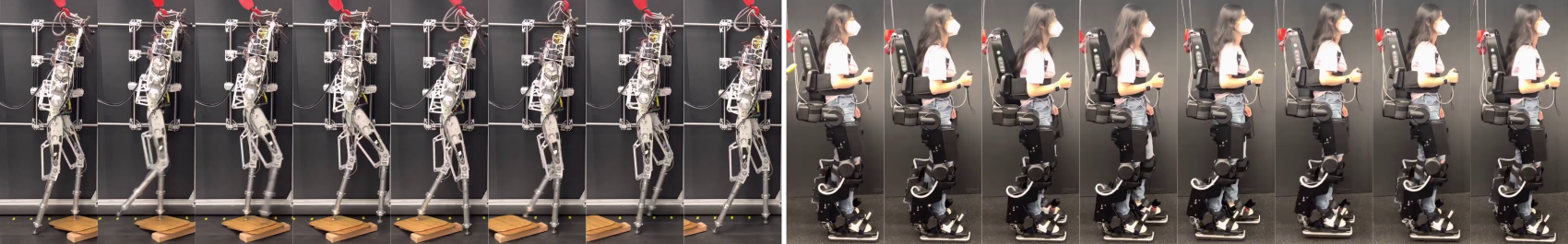}
    \caption{Gait tiles demonstrating robustness of the gaits generated with both torque and the extended saltation matrix in the cost function ($w_1,w_2 > 0$).}
    \label{fig: robustness}
    \vspace{-4mm}
\end{figure*}

\begin{figure}
    \centering
    \includegraphics[width=\linewidth]{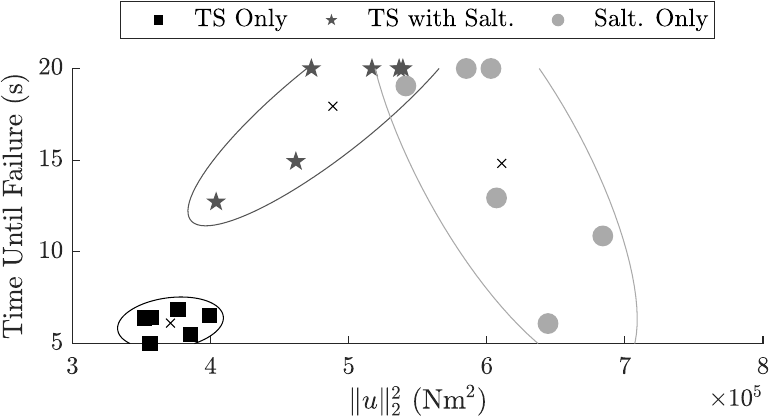}
    \caption{Simulation results on the Atalante exoskeleton for 6 random exoskeleton subject models. Each marker indicates an individual simulation with the corresponding time until failure (defined as the COM height falling below 0.4 meters) and nominal torque-squared evaluation. Each simulation was limited to 20 seconds total. The results for each gait condition are highlighted by ellipses constructed using 2-sigma fits to the data.}
    \label{fig:exo_sim}
    \vspace{-2mm}
\end{figure}

\newsubsec{AMBER-3M Simulation Results}
The three gaits were first demonstrated in a planar RaiSim \cite{raisim} simulation environment with randomly generated terrain. 
As shown in Fig. \ref{fig: ambersimgaits}, both gaits generated with the inclusion of the saltation matrix ($w_2 = 1$) were able to walk on rough terrain, while the gait generated with only torque-squared ($w_2 = 0$) failed. The figure also compares three environmental guard conditions: flat ground (as captured by the nominal guard condition), a 1deg slope, and a 2cm step. As illustrated by the phase portraits of the zero dynamics (selected as linearized forward hip position and velocity), the gaits with $w_2 = 1$ again show improved robustness. However, it is interesting to note that the gait with $w_1 = 0$ suffers from significantly increased torque while the gait with both terms ($w_1,w_2 = 1$) only has a moderate increase in required torque.

\newsubsec{AMBER-3M Experimental Results}
Once demonstrated in simulation, the three gaits were also demonstrated on hardware, as shown in Fig. \ref{fig:gaittiles}. The generated joint-level trajectories are enforced on AMBER-3M using an off-board joint-level PD controller 
that computes desired torques and sends them to the on-board motor controllers via UDP communication. The motor driver communication and control logic run at approximately 1kHz.
As shown in Fig. \ref{fig:gaittiles}, the gait generated using only torque-squared ($w_2 = 0$), and the gait generated using only the extended saltation matrix ($w_1 = 0$) were unstable, while the gait generated with the inclusion of both torque-squared and the extended saltation matrix ($w_1,w_2 > 0$) in the cost function was independently stable. 

To further demonstrate the performance of the gait generated with $w_1,w_2 > 0$, several robustness tests were also performed, as shown in Fig. \ref{fig: robustness}. For these experiments, random wooden objects were placed on the treadmill in front of AMBER-3M. These experiments highlight the robustness of the gait generated with the inclusion of both torque-squared and the extended saltation matrix in the cost function.

\newsec{Exoskeleton Results}
The second robotic platform used in this work is the Atalante lower-body exoskeleton -- a 3D bipedal platform capable of realizing crutch-less exoskeleton locomotion for people with complete motor paraplegia \cite{harib2018feedback}. In these experiments, sets of three gaits were generated for the exoskeleton with various human-subject models, as well as for the empty exoskeleton. As before, all constraints and bounds of the gait generation framework were held constant except for $w_1$ and $w_2$, and the generated joint-level trajectories were enforced using PD control.

The Atalante exoskeleton has 12 motorized joints (3 controlling the spherical position of each hip, 1 for each knee, and 2 in each ankle). As with AMBER, the outputs are again selected as the positions of the motorized joints, i.e. $y^a(q):= q^a \in \R^{12}$. 
%
Since the Atalante exoskeleton is fully-actuated when one foot is constrained to remain flat against the ground, gaits are generated using the \textit{Partial} Hybrid Zero Dynamics method \cite{ames2014human}, an extension of the HZD method. These gaits are described using $7^{\text{th}}$-order B\'ezier polynomials $(\alpha \in \R^{12 \times 8})$ and are parameterized using time as the phasing variable. It is important to note that by using time as a phase-variable, the theoretical guarantees of stability are no longer valid. For this reason, many results using periodic orbits on 3D robots also incorporate regulators to stabilize the walking \cite{reher2020algorithmic}. However, in this work, since we are interested in how \textit{robust} the periodic gaits are independent of regulators, no additional regulators were used (aside from desired output filtering to prevent discontinuities caused by early impacts). The generated joint-level trajectories are enforced on the Atalante exoskeleton using an on-board PD controller which sends current commands to the low-level motor drivers. 

\newsubsec{Atalante Simulation Results}
First, sets of three gaits were generated and deployed for six human models in a 3D simulation environment. As with prior work \cite{gurriet2018towards,gurriet2019towards, tucker2020preference, tucker2020human}, a human-exoskeleton model is synthesized by collecting the human's height, mass, thigh length, and shank length and using this information to approximate the remaining human segment inertias and remaining segment lengths based on anthropomorphic models from \cite{winter2009biomechanics}.

The simulation results, illustrated in Fig. \ref{fig:exo_sim}, found that the gaits with $w_2 > 0$ resulted in more stable steps being taken before the exoskeleton fell (characterized by the COM vertical height falling below 0.4 meters). However, the gait generated with only the saltation matrix resulted in significantly increased torque. In contrast, using both the saltation matrix and torque-squared in the cost function resulted in increased robustness with only a small increase in torque.

\newsubsec{Atalante Experimental Results}
Once demonstrated in simulation, a set of three gaits was also generated for the empty Atalante exoskeleton and deployed on hardware. The motivation for conducting experiments with the empty exoskeleton is to isolate the effects of the generated gaits independent of the human-subject's motion inside of the exoskeleton. The performance of the generated gaits was evaluated by whether or not the exoskeleton could locomote without operator interference for 3 meters. As shown in Fig. \ref{fig:gaittiles}, only the gait generated with both torque-squared and the saltation matrix ($w_1, w_2 > 0$) was able to successfully and independently walk for the full 3-meter test. 

Lastly, a set of gaits was also generated and deployed on the Atalante exoskeleton for a human subject. Again, only the gait generated with the inclusion of both cost function terms ($w_1, w_2 > 0$) was able to successfully complete the 3-meter walk test. These results are shown in Fig. \ref{fig: robustness} and in the supplemental video \cite{video}.

\section{Conclusion}

This work explored the application of saltation matrices for the generation of robust periodic walking gaits. 
%
%
The results found that by jointly minimizing the norm of the extended saltation matrix and torque, nominal gaits generated using the HZD framework successfully yielded robust locomotion.
It is interesting to note that on both platforms (AMBER-3M and the Atalante exoskeleton), gaits generated using only the saltation matrix performed well in simulation, but failed on hardware. This is most likely due to the fact that the saltation only gait result in much higher commanded torques, which leads to undesirable performance on hardware. Thus, future work includes exploring how to systematically balance this performance-robustness trade-off. 
Ultimately, the results of our work further expands our understanding of the metrics underlying \textit{robust} robotic walking such that in the future, stable and robust nominal gaits can be successfully generated without the need for extensive constraint tuning.


\small{
\section*{ACKNOWLEDGMENTS}
The authors would like to thank the entire Wandercraft team for their continued guidance and technical support with Atalante.}

\newpage 
\bibliographystyle{IEEEtran}
\balance
\bibliography{./Bibliography/IEEEabrv, ./Bibliography/References}

\clearpage
\clearpage

\end{document}